\documentclass[runningheads]{llncs}

\usepackage{eccv}

\usepackage{eccvabbrv}

\usepackage{graphicx}
\usepackage{booktabs}
\usepackage{siunitx}
\usepackage{pifont}
\usepackage{xcolor}
\usepackage[accsupp]{axessibility}  

\usepackage{hyperref}

\usepackage{orcidlink}
\usepackage{multirow}
\usepackage{array}

\begin{document}

\title{Invoice Haystack: Benchmarking Document Retrieval and Visual Question Answering Under Strong Visual Homogeneity} 

\titlerunning{Invoice Haystack} \author{ Heethanjan Kanagalingam\inst{1}\orcidlink{0009-0005-0869-5235} \and Thenukan Pathmanathan\inst{2}\orcidlink{0009-0008-2495-7903} \and Mokeeshan Vathanakumar\inst{1}\orcidlink{0009-0003-4415-8336} \and Basim Azam\inst{1}\orcidlink{0000-0002-3367-6467} \and Sarah Monazam Erfani\inst{1}\orcidlink{0000-0003-0885-0643} \and Naveed Akhtar\inst{1}\orcidlink{0000-0003-3406-673X} } \authorrunning{Heethanjan et al.} \institute{ The University of Melbourne, Melbourne, Australia \and Lakehead University, Thunder Bay, Canada\\ \email{heethanjanheetha@gmail.com} }

\maketitle

\begin{abstract}
Vision Language Models have achieved near-human performance on single-document Visual Question Answering, yet their effectiveness degrades significantly when retrieving information from large collections of visually homogeneous documents. Existing multi-document benchmarks aggregate diverse document types, creating artificial separation in embedding space that does not reflect enterprise document repositories where thousands of records share identical visual templates. We identify this as embedding collapse and introduce Invoice Haystack, a benchmark with  1,500 anonymized invoice images paired with 200 discriminative question-answer pairs, specifically designed to stress-test retrieval under strong visual homogeneity. Invoice Haystack exhibits a mean pairwise cosine similarity of  0.73, compared to 0.38 (DocHaystack) and 0.31 (InfoHaystack) in existing benchmarks, posing a fundamentally more challenging retrieval problem. Addressing the identified challenge, we propose VL-RAG, a hybrid retrieval-augmented generation framework that jointly leverages text and visual embeddings to harness the complementary strengths of both modalities, followed by a VLM-based verification filter for precise document identification.  VL-RAG achieves 60.0\% Recall@1  on Invoice Haystack-500, outperforming existing state-of-the-art method by up to an absolute 13.5 percentage points. It further improves retrieval considerably on DocHaystack-1000 (77.1\% vs.\ 75.2\%) and InfoHaystack-1000 (84.5\% vs.\ 80.0\%), establishing the proposed dual-stream fusion as a consistently superior retrieval strategy across both homogeneous and heterogeneous document collections.

\textit{The Invoice Haystack benchmark dataset, code, and project details are available at https://heethanjan.github.io/invoice-haystack/.}


\end{abstract}

\begin{figure} [tb]
    \centering
    \includegraphics[width=1\linewidth]{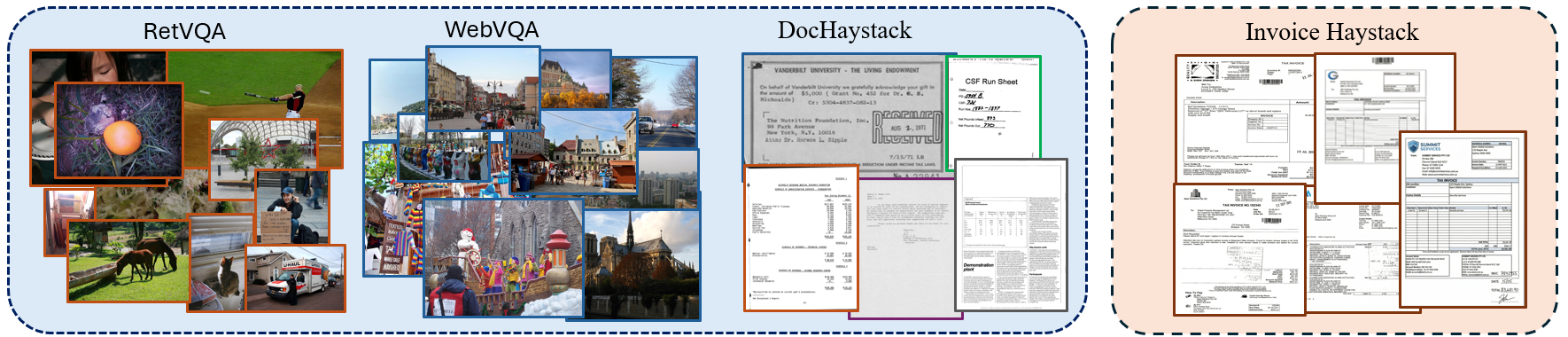}
    \caption{Comparison of data domains across visual retrieval benchmarks. Unlike general visual question answering datasets like RetVQA \cite{penamakuri2023answer}, and WebVQA \cite{webqa}, which rely on natural imagery, or Document Haystack \cite{doc_haystack_cvpr}, which aggregates diverse text documents, our proposed Invoice Haystack (right) targets the specialized domain of financial documentation. This focused scope challenges models to distinguish between visually similar transactional forms, requiring a deeper understanding of dense tabular structures and spatial layouts compared to general image or document collections.}
    \vspace{2mm}
    \label{fig:datasets}
\end{figure}

\section{Introduction}

The rapid advancement of Vision Language Models (VLMs) has fundamentally transformed the landscape of Document AI, catalyzing a paradigm shift from traditional modular pipelines comprising Optical Character Recognition (OCR), layout analysis, and rule-based information extraction \cite{xu2020layoutlm, LayoutLMv2, LayoutLMv3} toward unified end-to-end architectures capable of simultaneous visual perception and semantic understanding \cite{donut, pix2struct}. Contemporary state-of-the-art VLMs, including  ChatGPT-5.2 \cite{gpt-5.2},  Gemini 3.0 Pro \cite{gemini_3}, Claude 3.7 Sonnet \cite{claude35}, and Qwen3-VL \cite{bai2025qwen3vltechnicalreport}, have demonstrated unprecedented proficiency in single-document Visual Question Answering (VQA), achieving near-human performance on established benchmarks such as DocVQA \cite{docvqa}, InfographicVQA \cite{infographicvqa}, ChartQA \cite{chartqa}, and VisualMRC \cite{visualmrc}. These models leverage transformer-based architectures with cross-modal attention mechanisms \cite{attention_is_all, vilbert, clip} to jointly encode visual features and textual content, effectively solving the single-document comprehension problem for many practical applications \cite{llava, llava15}. However, these models still face significant challenges when tasked with reasoning over extensive collections of images or documents \cite{wu2025visualhaystacksvisioncentricneedleinahaystack}, limiting their deployment in real-world applications such as financial auditing, legal discovery, and supply chain management due to the lack of dedicated benchmarks for evaluating multi-document retrieval and reasoning performance \cite{kashyap2025structtextsynthetictabletotextapproach}.

The existing multi-image retrieving and reasoning benchmarks are primarily constructed on a small scale, as highlighted in works such as \cite{kazemi2024remidatasetreasoningmultiple, du2025easyhardmirbenchmark}. Each question in these benchmarks is paired with only up to 30 images, a limited scope that does not align with real-world scenarios requiring retrieval across thousands of files. Recent efforts, such as the DocHaystack and InfoHaystack benchmarks \cite{doc_haystack_cvpr}, have attempted to address this by scaling contexts up to 1,000 documents. However, while these benchmarks successfully address the challenges of scale and ambiguity, they fail to account for \textit{visual homogeneity}. Datasets like DocHaystack rely on visually diverse collections, ranging from bar charts to handwritten letters, that possess distinct visual features. Consequently, in the high-dimensional embedding space, these data points map to distinct clusters, artificially inflating retrieval performance because the negatives are easily distinguishable. This setup does not reflect operational enterprise realities, where a repository of thousands of invoices may utilise only a dozen templates. 

To rigorously address this gap, we introduce the Invoice Haystack benchmark. We identify invoices as the ideal stress test, as they combine complex spatial structures with sparse, high-value information that is visually repetitive but semantically distinct. The benchmark is constructed through a rigorous four-stage pipeline comprising anonymization of personally identifiable information (PII), VLM-based question generation, automated LLM filtering, and expert human validation and correction.

To enable VLMs to navigate this challenging environment, we propose VL-RAG (Vision-Language RAG). We posit that in homogeneous haystacks, neither visual features nor textual features alone are sufficient for differentiation. Vision encoders collapse visually identical templates into indistinguishable representations, a phenomenon we term embedding collapse, while text-only approaches discard valuable layout and structural signals~\cite{shen-etal-2022-vila}. VL-RAG fundamentally restructures the retrieval hierarchy by incorporating both vision and text encoders to capture dense embeddings. By integrating these semantic signals alongside visual features, VL-RAG enables precise and discriminative document retrieval across visually homogeneous collections.

Our key contributions are: \textbf{(1)  Invoice Haystack Benchmark.} We introduce the Invoice Haystack benchmark of 1,500 anonymized invoice images paired with 200 discriminative question-answer pairs, specifically engineered to stress-test retrieval under extreme visual homogeneity. The benchmark is evaluated across three corpus scales, 500, 1,000, and 1,500 documents, enabling controlled measurement of how retrieval performance degrades as the size of the homogeneous pool increases. Invoice Haystack exhibits a mean pairwise cosine similarity of 0.73,  nearly double the 0.31--0.38 range of prior benchmarks, posing a qualitatively harder retrieval challenge than any existing multi-document dataset. \textbf{(2) VL-RAG Method.} We propose a hybrid dual-stream approach that integrates dense text embeddings (BGE-Large) alongside structural visual features (SigLIP, OpenCLIP) to address the embedding collapse that cripples vision-only retrieval in homogeneous collections. Proposed VL-RAG achieves up to 13.5 percentage points gain on Invoice Haystack for Recall@1, while maintaining across-the-board superiority over existing state-of-the-art methods for Document Haystack and InfoHaystack for Recall@1,3,5. Our results conclusively establish our dual-stream fusion technique as the priority method. 

\section{Related Work}

\label{sec:related}

\smallskip\noindent\textbf{Document Understanding with Vision-Language Models.} Document AI has evolved from modular pipelines combining OCR \cite{tesseract, abbyy_ocr}, layout analysis \cite{layout_detection, table_detection}, and NLP models \cite{bert, roberta} toward unified end-to-end architectures. LayoutLM \cite{xu2020layoutlm, LayoutLMv2, LayoutLMv3} pioneered spatial-aware pre-training by integrating 2D position embeddings with text, while OCR-free models including Doughnut \cite{donut}, Pix2Struct \cite{pix2struct}, and Nougat \cite{nougat} eliminated external OCR dependency through vision transformers \cite{swin_transformer, vit}. Recent multimodal architectures like mPLUG-DocOwl \cite{mplug_docowl}, UniDoc \cite{unidoc}, and UDOP \cite{udop} unified vision, text, and layout representations. Contemporary VLMs including ChatGPT-5.2 \cite{gpt-5.2}, Gemini 3.1 Pro \cite{gemini_3.1}, Qwen3-VL \cite{bai2025qwen3vltechnicalreport}, and InternVL-3.5\cite{wang2025internvl35advancingopensourcemultimodal} achieve over 90\% accuracy on single-document VQA benchmarks \cite{docvqa, infographicvqa, chartqa, visualmrc}. However, performance degrades substantially in multi-document settings. 
Notably, reasoning performance has been reported to degrade by up to 85\% as input length scales, even when models successfully retrieve the correct evidence \cite{du2025context}. This stark decline, despite the advent of million-token windows, indicates fundamental ongoing challenges in document-scale reasoning.

\smallskip\noindent\textbf{Multi-Document Benchmarks \&  Homogeneity Gap.} Early multi-document benchmarks including MultimodalQA \cite{multimodalqa} and WebQA \cite{webqa} limited contexts to 10-30 images per question \cite{penamakuri2023answer, fm2ds}, insufficient for enterprise-scale evaluation. DocHaystack and InfoHaystack \cite{doc_haystack_cvpr} recently advanced the field by scaling to 1,000 documents through multi-stage filtering with LLM-based question generation \cite{gpt-5.2, claude3} and human validation, ensuring answer uniqueness and visual grounding. Nevertheless, this design still falls short on reflecting real-world enterprise repositories, which are dominated by highly structured, heavily templated documents such as invoices, receipts, and standardised forms \cite{katti2018chargrid, xu2020layoutlm}. In these largely \textit{homogenous} collections, documents from different vendors or transactions often share near-identical visual layouts, leading to a collapse in visual embedding space. Yet, existing benchmarks do not evaluate this critical setting, leaving a fundamental gap in assessing retrieval performance reliably. 

\smallskip\noindent\textbf{Retrieval Methods \& Vision-only Limitations.} Contemporary vision-based retrieval leverages contrastive models pre-trained on natural images, including CLIP \cite{clip}, SigLIP \cite{siglip}, and DINOv2 \cite{dinov2}, combined with vision transformers \cite{vit, swin_transformer} and vision-language alignment modules \cite{blip2, instructblip}. ColPali \cite{colpali} introduced late interaction mechanisms for page-level retrieval, achieving state-of-the-art performance on visually rich document benchmarks. Text-based dense retrieval methods, including DPR \cite{dpr} and ColBERT \cite{colbert}, excel on text-based documents but remain limited when documents share strong visual and semantic similarity at scale. Hybrid approaches like Flamingo \cite{flamingo} and RA-CM3 \cite{racm3} combine modalities but focus on few-shot learning over natural image-text pairs rather than large-scale document retrieval.

\smallskip\noindent\textbf{Retrieval-Augmented Generation for Documents.} Text-based RAG systems combine retrieval with generation: RAG \cite{rag} integrates DPR \cite{dpr} and BART \cite{lewis2020bart}  for open-domain QA, REALM \cite{realm} incorporates retrieval into pre-training, Atlas \cite{atlas} demonstrates strong few-shot performance through Fusion-in-Decoder \cite{fid}, achieving 42.4\% exact match on NaturalQuestions \cite{kwiatkowski2019natural}  with only 64 training examples (45.1\% with a Wikipedia-only index), and recent advances including Self-RAG \cite{self_rag}, and CRAG \cite{crag} introduce retrieval-on-demand and corrective mechanisms. Multimodal RAG extensions remain limited: RA-CM3 \cite{racm3} evaluates on-web images, REVEAL \cite{reveal} uses Wikipedia images, and MuRAG \cite{murag} requires manual modality selection \cite{multimodal_rag, racm3}. V-RAG \cite{doc_haystack_cvpr} addresses visual document retrieval at scale through an ensemble of vision encoders (CLIP, SigLIP, and OpenCLIP) combined with a VLM-based relevance filtering module, achieving 9\% and 11\% improvement in Recall@1 over prior best baselines on the DocHaystack-1000 and InfoHaystack-1000 benchmarks, respectively, and enabling GPT-4o to improve by over 55\% on DocHaystack-200. However, purely vision-centric approaches remain limited when documents share strong visual similarity, as evidenced by the substantially lower retrieval performance of individual CLIP-based models on document-heavy collections compared to text-based BM25 retrieval using OCR \cite{robertson2009probabilistic}, motivating the need for hybrid architectures that jointly encode visual layout and textual content.


\section{Proposed Invoice Haystack Benchmark}

\begin{figure}[tb]
    \centering
    \includegraphics[width=1\linewidth]{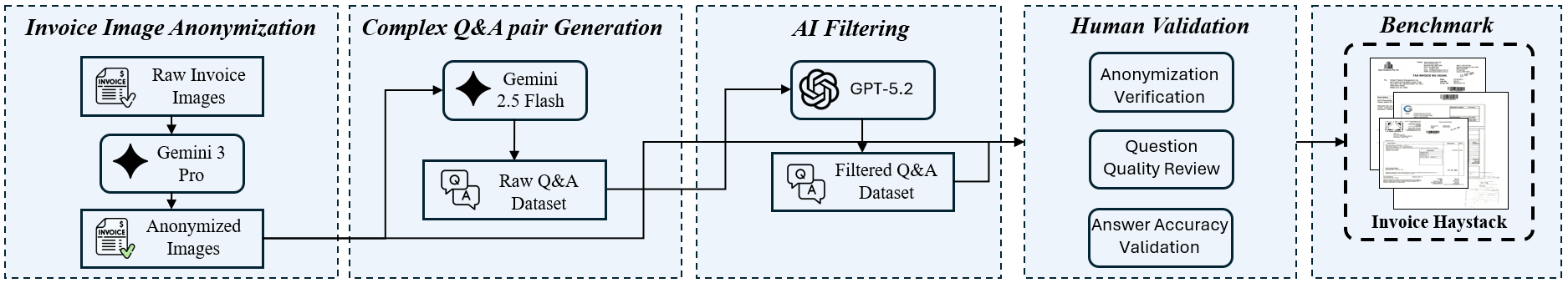}
    \caption{Invoice Haystack Generation Pipeline. Our dataset is curated through a comprehensive four-stage workflow. Step 1 (Invoice Image Anonymization) employs Gemini 3 Pro to sanitize raw invoice images, ensuring sensitive information is removed. Step 2 (Complex Q\&A Generation) utilizes Gemini 2.5 Flash to analyze the anonymized documents and synthesize a raw dataset of question-answer pairs. Step 3 (AI Filtering) refines this output using GPT-5.2 to filter for quality and relevance. Finally, Step 4 (Human Validation) subjects the filtered dataset to a three-pronged manual review—verifying anonymization, question quality, and answer accuracy—to produce the final Invoice Haystack benchmark.}
    \label{fig:pipeline}
\end{figure}

Our Invoice Haystack benchmark addresses a critical gap in document AI evaluation: assessing retrieval performance under strong visual homogeneity. Unlike existing benchmarks that aggregate diverse document types \cite{doc_haystack_cvpr, infographicvqa}, our benchmark targets template-heavy collections such as invoices, where documents follow a similar layout structure while differing only in fine-grained semantic content. We construct this benchmark through a four-stage pipeline combining multi-modal AI models with rigorous human oversight, as illustrated in Figure~\ref{fig:pipeline}.

\smallskip\noindent\textbf{Motivation.} Invoices provide an ideal domain for evaluating homogeneous document retrieval. They combine complex spatial layouts with multi-column tables \cite{xu2020layoutlm, LayoutLMv2}, critical information distributed across distinct regions (headers, line entries, footers), and extreme template standardization. In curating our corpus, we observed that a small number of recurring vendor templates accounted for the substantial majority of documents, with the remaining invoices distributed across a long tail of infrequent formats---a pattern consistent with known supplier-driven layout skew in enterprise invoice repositories \cite{KRIEGER2023200285}. Two invoices from the same template may exhibit cosine similarity exceeding 0.95 in vision-only embeddings yet contain completely different semantic content distinguishable only through text analysis. This structure directly mirrors real-world scenarios where systems must locate specific transactions among thousands of visually similar documents for auditing, compliance, and dispute resolution.


\subsection{Data Curation Pipeline}

The benchmark is constructed through a four-stage workflow - see  Figure~\ref{fig:pipeline}. The pipeline is designed to maximize data quality, preserve privacy, and ensure that every retained question-answer pair is both discriminative and visually grounded.

\smallskip\noindent\textbf{Stage 1: Invoice Image Anonymization.} We employ Gemini 3 Pro~\cite{gemini_3} to sanitize 2,000 raw invoice images via a structured multi-modal prompt targeting seven sensitive information categories: company and customer identifiers, physical addresses, contact details, corporate branding and logos, temporal identifiers such as invoice numbers and dates, personal identifiers such as account managers, and financial account details. Critically, all numerical values are preserved in their correct relational form — substituted figures are synthetically generated such that line totals, subtotals, tax computations, and final amounts remain mathematically consistent throughout, ensuring that anonymization does not corrupt the semantic integrity of the financial content. Layout geometry, including table structures, column alignments, fonts, and visual separators, is preserved exactly to prevent models from exploiting anonymization artifacts as discriminative signals in their predictions.

\smallskip\noindent\textbf{Stage 2: Complex Q\&A Generation.} Gemini 2.5 Flash~\cite{comanici2025gemini} analyzes each anonymized image to generate one question-answer pair per document, yielding 2,000 raw candidate pairs. Question design enforces unique document identification through specific visual entities, such as company names, customer identifiers, product descriptions, or transaction dates, with multi-field questions (e.g., \textit{"What is the order no associated with the contract invoice transaction dated 15-OCT-2023 and building no 506890?"}) preventing non-unique answers across template instances and ensuring retrieval requires precise semantic matching. Generic questions (e.g., \textit{"What is the invoice total?"}) are explicitly excluded. Answers are constrained to short extractive spans (1–2 words preferred, 5-word maximum) appearing verbatim in the image, with generation temperature set to 0.7 to balance output diversity and format adherence.

\smallskip\noindent\textbf{Stage 3: AI Filtering.} GPT-5.2 \cite{gpt-5.2} performs automated quality filtering over the 2,000 candidate pairs, removing questions that are generic, ambiguous, non-discriminative, or answerable without image access. This reduces the dataset to 200 high-quality pairs that satisfy uniqueness and visual dependency constraints. The filtering step acts as a scalable quality gate, efficiently eliminating the majority of low-quality outputs before human review.

\smallskip\noindent\textbf{Stage 4: Human Validation.} Expert annotators apply a three-pronged manual review to all 200 filtered question-answer pairs and their associated images. Reviewers verify anonymization completeness by scanning for residual PII, including real company names, addresses, contact information, invoice numbers enabling external correlation, and recognizable branding. Question quality is assessed for discriminative power, clarity, and non-trivial reasoning requirements. Answer accuracy is confirmed by direct visual inspection of source images, ensuring that answers appear exactly as specified and no ambiguous alternatives exist. Entries failing any criterion are corrected or changed. Additionally, the full corpus of 2,000 anonymized images underwent manual inspection to remove documents with anonymization failures, rendering artifacts, or quality issues, reducing the final corpus to 1,500 high-quality invoice images. This produced the final validated benchmark of 200 question-answer pairs evaluated against 1,500 documents. Details of the prompts utilized across these stages, along with sample anonymized images and questions, are provided in the supplementary material.

\subsection{Benchmark Statistics and Analysis}

\begin{figure*}[tb]
    \centering
    \includegraphics[width=\linewidth]{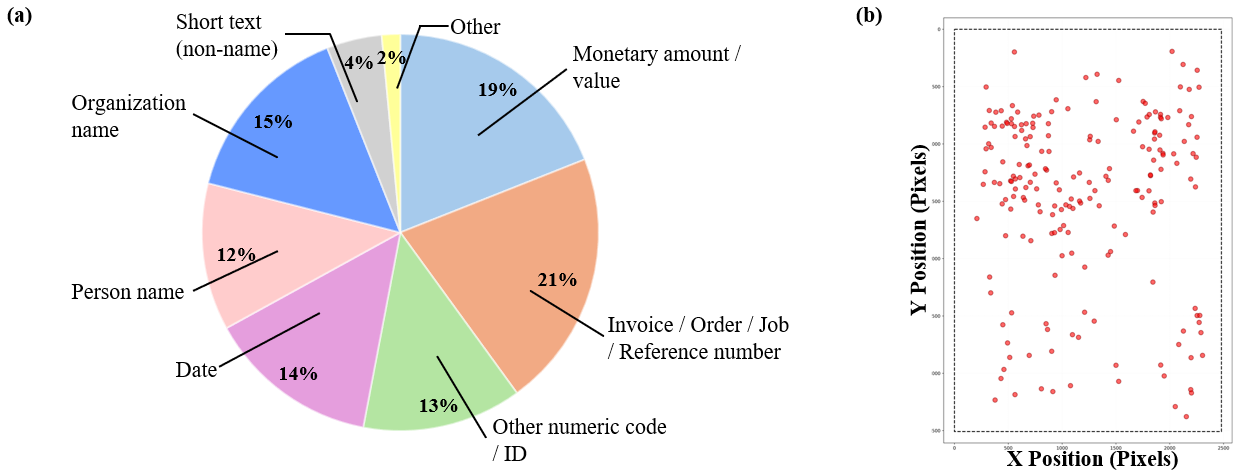}
    \caption{\textbf{(a)} Distribution of question types in Invoice Haystack across eight semantic categories. \textbf{(b)}  Spatial positioning of answers within the documents.}

    \label{fig:qa_distribution}
   
\end{figure*}


The final Invoice Haystack benchmark comprises 200 validated question-answer pairs evaluated against a corpus of 1,500 visually homogeneous invoice images. Figure~\ref{fig:qa_distribution}(a) presents the distribution of question types across the dataset. Questions span eight semantic categories targeting different invoice fields. This distribution reflects the diversity of information types present in real invoice documents and ensures comprehensive evaluation across different retrieval scenarios. The deliberate exclusion of yes/no questions and generic quantity queries enforces that every question demands specific document-grounded reasoning. Figure~\ref{fig:qa_distribution}(b) shows the spatial distribution of ground-truth answer locations across all invoice documents. Each point denotes the centre of the OCR bounding box matched to a ground-truth answer string, with all positions normalised to an A4 coordinate system (2,480 x 3,508 pixels). The clustering patterns reveal consistent yet varied spatial placement across diverse invoice layouts. This spatial structure confirms that answers are not trivially co-located and that retrieval models must reason over the full document rather than attending to a single fixed region.
Table~\ref{tab:benchmark_comparison} compares Invoice Haystack with the most closely related existing document retrieval benchmarks. Invoice Haystack exhibits a mean pairwise cosine similarity of 0.73, more than double the 0.31–0.38 range of prior benchmarks, reflecting a fundamentally harder retrieval setting in which documents cannot be easily distinguished by visual features alone.

\begin{figure*}[tb]
    \centering
    \includegraphics[width=\linewidth]{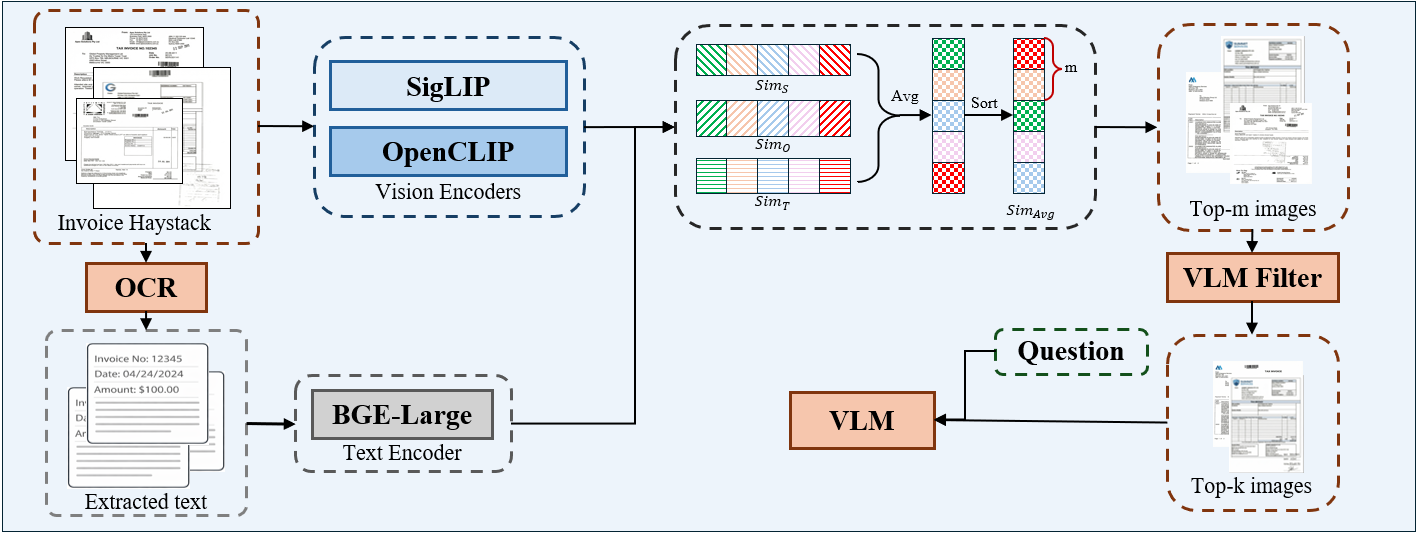}
    \caption{Proposed VL-RAG architecture. The approach operates via two parallel streams. The Vision Stream (top) employs SigLIP and OpenCLIP to capture layout and structural features, while the Text Stream (bottom) uses OCR followed by BGE-Large to encode semantic content. Similarity scores for all three encoders are averaged ($\text{Sim}_{\text{Avg}}$) to rank the corpus, and the top-$m$ candidates undergo VLM-based binary verification to produce the final top-$k$ output.}
    \label{fig:vlrag_arch}
\end{figure*}

\begin{table}[tb]
\centering
\caption{Comparison of Invoice Haystack with existing document retrieval benchmarks. Visual homogeneity is measured by mean within-category SigLIP cosine similarity. The proposed benchmark is high-potential and difficult and stronger in homogeneity.}
\label{tab:benchmark_comparison}
\begin{tabular}{lccc}
\hline
\textbf{Benchmark} & \textbf{Corpus Size} & \textbf{Doc Type} & \textbf{Homogeneity}\\
\hline
DocHaystack    & 1,000 & Mixed   & 0.38  \\
InfoHaystack   & 1,000 & Mixed   & 0.31  \\
\textbf{Invoice Haystack (Ours)}       & \textbf{1500} & \textbf{Invoices} & \textbf{0.73} \\
\hline
\end{tabular}

\end{table}

\section{Methodology: Proposed VL-RAG Approach}

The central hypothesis of this work is that in visually homogeneous document collections, vision-only and text-only embeddings are fundamentally insufficient for precise retrieval. Standard vision encoders trained on natural images tend to collapse template-based documents into indistinguishable representations, failing to resolve fine-grained semantic differences that exist only in textual content. To address this, we propose VL-RAG (Vision-Language Retrieval-Augmented Generation), a hybrid dual-stream framework that combines dense text embeddings with structural visual features. As illustrated in Figure~\ref{fig:vlrag_arch}, the proposed method operates through three core components: a text encoding stream that captures semantic content via OCR and dense embeddings, a vision stream that encodes layout and structural features, and a VLM-based verification filter that refines the final retrieval output. Together, these components enable precise document identification in homogeneous collections where visual similarity alone is insufficient.



\smallskip\noindent\textbf{Text Stream.} The text stream is designed to capture the high-precision semantic information that vision models systematically fail to capture. An OCR engine first extracts raw text from each invoice image, converting the visual document into a structured textual representation. This extracted text is then encoded using a dense text encoder, and the resulting embedding 
captures the precise semantic content of each document. Documents encoded in this manner are well-separated in text embedding space, provided the discriminative signal is absent in vision-only retrieval. The text stream exploits this separability, computing a similarity score $\text{Sim}_T$ for each query-document pair:
\begin{equation}
    \text{Sim}_T(q, d_i) = \frac{\phi_T(\hat{q}) \cdot \phi_T(t_i)}
    {\|\phi_T(\hat{q})\|\,\|\phi_T(t_i)\|},
    \label{eq:text_sim}
\end{equation}

where $\phi_T(\cdot)$ denotes the BGE-Large encoder, $t_i$ is the OCR-extracted text of document $d_i$, and $q$ represents the input query (with $\hat{q}$ being its textual representation).

\smallskip\noindent\textbf{Vision Stream.} The vision stream captures structural and layout-level features that complement the text signal. Following V-RAG~\cite{doc_haystack_cvpr}, we adopt SigLIP~\cite{siglip} and OpenCLIP~\cite{openclip} as the vision encoders, ensembling their outputs to form the visual component of VL-RAG. This choice is further validated by our ablation study (Table~\ref{tab:ablation}), which shows that the SigLIP~+~OpenCLIP combination consistently outperforms any single vision encoder. Together, these encoders compute similarity scores $\text{Sim}_S$ and $\text{Sim}_O$ respectively, as follows:

\begin{equation}
    \text{Sim}_S(q, d_i) = \frac{\phi_S(I_q) \cdot \phi_S(I_i)}
    {\|\phi_S(I_q)\|\,\|\phi_S(I_i)\|}, \quad
    \text{Sim}_O(q, d_i) = \frac{\phi_O(I_q) \cdot \phi_O(I_i)}
    {\|\phi_O(I_q)\|\,\|\phi_O(I_i)\|}
    \label{eq:vision_sim}
\end{equation}
where $\phi_S(\cdot)$ and $\phi_O(\cdot)$ denote the SigLIP and OpenCLIP encoders respectively, and $I_i$ is the image of document $d_i$.


\smallskip\noindent\textbf{Score Fusion.} As depicted in Figure~\ref{fig:vlrag_arch}, the three similarity scores from the text stream ($\text{Sim}_T$) and vision stream ($\text{Sim}_S$, $\text{Sim}_O$) are combined through average fusion to produce a unified ranking score:
\begin{equation}
    \text{Sim}_{\text{Avg}} = \frac{1}{3}\left(\text{Sim}_T + \text{Sim}_S + \text{Sim}_O\right).
\end{equation}
The corpus is then sorted by $\text{Sim}_{\text{Avg}}$ in descending order. The top-$m$ candidate set is formally defined as:
\begin{equation}
    \mathcal{C}_m = \underset{\mathcal{S} \subseteq \mathcal{D},\,|\mathcal{S}|=m}{\arg\,\text{top}} \;\text{Sim}_{\text{Avg}}(q, d_i).
    \label{eq:candidate_set}
\end{equation}
This fusion ensures that a document is promoted only when it scores well across both modalities, reducing false positives caused by either visual layout matches that lack textual alignment or textual matches in documents with divergent layouts. The top-$m$ candidate set $\mathcal C_m$ is further filtered as below. 

\smallskip\noindent\textbf{VLM Filter.} Vector similarity scoring provides a high-recall shortlist but does not guarantee that any retrieved document actually answers the query. To convert this high-recall set into a high-precision result, VL-RAG applies a VLM Filter stage over the top-$m$ candidates. Each candidate image $I_i$ is paired with the original query $q$ and passed to Qwen3-VL, which performs binary verification via the prompt: \textit{``[question] Can this image provide the answer to this question? Answer only yes or no.''}. The model's response is parsed, and only candidates that return a \texttt{yes} response are retained:
\begin{equation}
    \mathcal{R}_k = \{d_i \in \mathcal{C}_m \mid \text{VLM}(I_i, q) = \texttt{yes}\}
    \label{eq:vlm_filter}
\end{equation}
If no candidate passes verification, the top-$m$ shortlist is returned unfiltered as a fallback to ensure that a result is always produced. This stage is essential in homogeneous settings where multiple visually similar documents score highly under vector similarity. The VLM provides the semantic reasoning capacity to resolve such ambiguity that vector similarity alone cannot.

\section{Experiments}

\smallskip\noindent\textbf{Implementation Details.} All of our experiments are conducted under a standardized retrieval-augmented generation pipeline to ensure a fair comparison. For the vision stream, we use frozen pre-trained checkpoints of SigLIP~\cite{siglip} ViT-SO400M/14@384 variant and OpenCLIP~\cite{openclip} ConvNeXt-XXL@1024 variant, evaluated in zero-shot mode without fine-tuning. For the text stream, we employ BGE-Large-En-v1.5~\cite{xiao2024c} as the dense text encoder, applied over text extracted via a DeepSeek OCR engine. The VLM filter and final answer generation both use Qwen3-VL-8B Instruct~\cite{bai2025qwen3vltechnicalreport}, configured with greedy decoding and a maximum of 128 new tokens. For the VLM filter, we set the candidate pool size to $m = 10$ at $k \in \{1, 3, 5\}$ for all retrieval experiments (ablation for the selection of $m$ is provided in the supplementary material) and report results.  All inference is executed on NVIDIA A100 (80GB) GPUs. For each query, the haystack is constructed by combining the ground-truth document with $N-1$ distractors drawn from the respective benchmark split.

\smallskip\noindent\textbf{Baselines.} We compare VL-RAG against a range of retrieval approaches spanning sparse text, dense vision, and vision-language methods. BM25 \cite{robertson2009probabilistic} serves as the sparse text baseline applied over OCR-extracted content. Vision-only baselines include CLIP~\cite{clip} ViT-L/14@336 variant, SigLIP~\cite{siglip} ViT-SO400M/14@384 variant, OpenCLIP~\cite{openclip} ConvNeXt-XXL@1024 variant, Jina-CLIP-v2~\cite{koukounas2024jina}, and Nomic-Embed-Vision-v1.5~\cite{nussbaum2024nomic}. The primary multimodal baseline is V-RAG~\cite{doc_haystack_cvpr}, a recent vision-centric retrieval-augmented generation system. For a fair comparison, we adapt V-RAG to use Qwen3-VL-8B-Instruct~\cite{bai2025qwen3vltechnicalreport} as its VLM component, matching the VLM configuration in VL-RAG. All models are evaluated under identical conditions across three metrics: Recall@1, Recall@3, and Recall@5.

\subsection{Main Results}

Table~\ref{tab:main_results} reports Recall@1, Recall@3, and Recall@5 retrieval performance across all three benchmarks. We organize findings by benchmark domain to highlight the distinct behavior of each model class under visual homogeneity.

\smallskip\noindent\textbf{Invoice Haystack:} This benchmark represents the hardest retrieval setting due to extreme visual homogeneity. Vision-only models exhibit severe performance collapse across all corpus scales. CLIP achieves only 7.0\% Recall@1 on the 1500-document split, and Jina-CLIP-v2 degrades further to 2.5\%, performing near chance. SigLIP and OpenCLIP fare marginally better at 20.0\% and 19.5\%, respectively, yet remain far below practical utility. Even V-RAG, which aggregates multiple vision encoders, reaches only 40.0\% Recall@1 at the largest scale. This degradation reflects embedding collapse: as visually identical templates proliferate, vision encoders cannot resolve instance-level semantic differences. Notably, BM25 achieves 38.5\% Recall@1 on the 1500-split, outperforming all vision-only baselines and only 1.5\% less than V-RAG itself, which directly demonstrates that textual content is a primary discriminative signal in this domain. VL-RAG achieves 50.0\% Recall@1 on the hardest split, improving over V-RAG by 10 percentage points and confirming the necessity of explicit text encoding in visually homogeneous retrieval.

\smallskip\noindent\textbf{Document Haystack.} On this visually heterogeneous benchmark, all models perform substantially better, and the performance ordering shifts. V-RAG achieves 75.2\% Recall@1 at the 1000-document scale, while Nomic-Embed-Vision performs competitively at 61.5\%. VL-RAG reaches \textbf{77.1\%} Recall@1 at 1000 documents, improving upon V-RAG across all scales. The smaller absolute gain relative to Invoice Haystack confirms that visual features are already sufficiently discriminative in heterogeneous corpora, and the text stream provides a complementary rather than compensatory signal.

\smallskip\noindent\textbf{InfoHaystack.} Results follow a similar trend. V-RAG achieves 80.0\% Recall@1 at the 1000-document scale, while VL-RAG reaches \textbf{84.5\%}, a gain of 4.5 percentage points. The most pronounced improvement occurs at the 200-document scale, where VL-RAG surpasses V-RAG by 5.8 percentage points (94.2\% vs 88.4\%). CLIP performs more competitively in this domain than on Invoice Haystack, consistent with its natural image pre-training aligning better with visually diverse infographic content.

\begin{table*}[t]
\centering
\caption{Retrieval performance  (Recall@1 / Recall@3 / Recall@5) across Document Haystack, InfoHaystack, and proposed Invoice Haystack benchmarks. Best results per column are {bold}. Proposed VL-RAG achieves across-the-board superior performance.}
\label{tab:main_results}
\resizebox{\textwidth}{!}{%
\begin{tabular}{l|ccc|ccc|ccc|ccc|ccc|ccc|ccc|ccc|ccc}
\hline
 & \multicolumn{9}{c|}{\textbf{Document Haystack}} & \multicolumn{9}{c|}{\textbf{InfoHaystack}} & \multicolumn{9}{c}{\textbf{Invoice Haystack}} \\
 & \multicolumn{3}{c|}{100} & \multicolumn{3}{c|}{200} & \multicolumn{3}{c|}{1000} & \multicolumn{3}{c|}{100} & \multicolumn{3}{c|}{200} & \multicolumn{3}{c|}{1000} & \multicolumn{3}{c|}{500} & \multicolumn{3}{c|}{1000} & \multicolumn{3}{c}{1500} \\
\textbf{Model} & R@1 & R@3 & R@5 & R@1 & R@3 & R@5 & R@1 & R@3 & R@5 & R@1 & R@3 & R@5 & R@1 & R@3 & R@5 & R@1 & R@3 & R@5 & R@1 & R@3 & R@5 & R@1 & R@3 & R@5 & R@1 & R@3 & R@5 \\
\hline
BM25 (OCR)         & 69.7 & 79.8 & 81.7 & 65.1 & 75.2 & 78.0 & 62.4 & 70.6 & 73.4 & 69.7 & 79.8 & 81.7 & 47.7 & 63.2 & 69.7 & 32.9 & 43.2 & 51.0 & 43.0 & 55.5 & 58.0 & 41.5 & 48.0 & 52.0 & 38.5 & 45.5 & 49.5 \\
Jina-CLIP-v2        & 62.4 & 77.1 & 84.4 & 58.7 & 69.7 & 78.0 & 35.8 & 54.1 & 61.5 & 62.4 & 77.1 & 84.4 & 62.6 & 84.5 & 88.4 & 38.7 & 59.4 & 67.7 & 4.5  & 9.5  & 12.5 & 4.0  & 5.5  & 8.0  & 2.5  & 5.0  & 5.0  \\
Nomic-Embed        & 78.9 & 84.4 & 87.2 & 78.0 & 83.5 & 86.2 & 61.5 & 69.7 & 74.3 & 78.9 & 84.4 & 87.2 & 54.2 & 61.9 & 63.2 & 49.7 & 59.4 & 60.0 & 34.5 & 44.5 & 49.0 & 28.0 & 38.0 & 44.0 & 26.5 & 36.0 & 40.0 \\
CLIP               & 46.8 & 66.1 & 68.8 & 44.0 & 64.2 & 66.1 & 27.5 & 41.3 & 47.7 & 46.8 & 66.1 & 68.8 & 66.5 & 78.7 & 85.8 & 53.6 & 68.4 & 72.3 & 10.0 & 16.0 & 22.0 & 8.0  & 13.0 & 15.0 & 7.0  & 11.0 & 14.0 \\
SigLIP             & 57.8 & 66.1 & 72.5 & 55.1 & 60.6 & 64.2 & 41.3 & 51.4 & 54.1 & 57.8 & 66.1 & 72.5 & 65.8 & 83.2 & 89.0 & 45.8 & 62.6 & 69.7 & 28.0 & 40.5 & 46.0 & 24.0 & 33.0 & 39.5 & 20.0 & 30.0 & 37.0 \\
OpenCLIP           & 66.1 & 75.2 & 81.7 & 56.9 & 72.5 & 77.1 & 41.3 & 59.6 & 66.1 & 66.1 & 75.2 & 81.7 & 77.4 & 87.7 & 89.7 & 57.4 & 72.3 & 79.4 & 22.0 & 31.5 & 37.0 & 20.0 & 28.0 & 31.5 & 19.5 & 24.0 & 28.0 \\
V-RAG             & 88.1 & 89.9 & 89.9 & 83.5 & 87.2 & 87.2 & 75.2 & 80.7 & 80.7 & 92.3 & 95.5 & 95.5 & 88.4 & 92.9 & 92.9 & 80.0 & 83.2 & 85.2 & 46.5 & 49.0 & 49.0 & 43.0 & 45.5 & 46.0 & 40.0 & 43.0 & 43.0 \\
\textbf{VL-RAG (Ours)} & \textbf{89.9} & \textbf{91.7} & \textbf{91.7} & \textbf{86.2} & \textbf{88.1} & \textbf{88.1} & \textbf{77.1} & \textbf{81.7} & \textbf{81.7} & \textbf{94.2} & \textbf{96.8} & \textbf{96.8} & \textbf{94.2} & \textbf{96.8} & \textbf{96.8} & \textbf{84.5} & \textbf{87.7} & \textbf{89.0} & \textbf{60.0} & \textbf{63.5} & \textbf{63.5} & \textbf{53.0} & \textbf{57.0} & \textbf{57.5} & \textbf{50.0} & \textbf{53.5} & \textbf{54.0} \\
\hline
\end{tabular}}

\end{table*}

\smallskip\noindent\textbf{Visual Question Answering (VQA) results.}
VQA accuracy had been evaluated across three frameworks, which are zero-shot, V-RAG, and our proposed VL-RAG. With each V-RAG and VL-RAG entry reflecting the best accuracy achieved across Top-1/3/5 retrieval depths(Table~\ref{tab:vlm_comparison_filtered}). To evaluate these benchmarks, we employ a model-based assessment (LLM as a Judge) by leveraging GPT-4o-mini \cite{hurst2024gpt} following \cite{doc_haystack_cvpr}. Zero-shot results are included only for smaller corpus scales (100 and 200 documents for DocHaystack and InfoHaystack; 500 for Proposed Invoice Haystack for capable models), as the larger splits, and even the 500-document Invoice split for InternVL3-8B and Qwen3-VL-8B, exceed the practical context-window capacities. These zero-shot numbers serve as a meaningful lower bound: even frontier models score only 18.0\% (Gemini) and 33.0\% (GPT-5.2) on InvoiceVQA-500 without retrieval support, confirming that the proposed Invoice Haystack poses a challenge that cannot be bypassed by simply expanding context. Across all four VLMs and corpus scales, VL-RAG consistently outperforms V-RAG with gains ranging from 2 to 16 percentage points. The improvements are most pronounced on our Invoice Haystack benchmark, where visual homogeneity is most severe: Gemini-3-Flash improves from 51.0\% to \textbf{60.0\%} on InvoiceVQA-500 and from 44.0\% to \textbf{53.0\%} on InvoiceVQA-1500, while GPT-5.2 rises from 49.5\% to \textbf{61.0\%} at the 500-document scale. Among all models, Qwen3-VL-8B achieves the highest single accuracy of \textbf{63.5\%} on InvoiceVQA-500.

On the heterogeneous benchmarks, VL-RAG continues to improve over V-RAG, though with smaller margins consistent with visual features already providing reasonable discriminability in those settings. Gemini-3-Flash reaches \textbf{76.1\%} on DocHaystack-1000 and \textbf{73.5\%} on InfoHaystack-1000 under VL-RAG. As expected, accuracy declines monotonically as corpus scale increases from 500 to 1500 documents across all models; however, VL-RAG maintains a clear and consistent advantage at every scale, demonstrating that incorporating textual signals alongside visual retrieval is a robust and universally beneficial strategy regardless of document homogeneity.
\begin{table*}[t]
\centering
\caption{VQA accuracy (\%) comparison across DocHaystack, InfoHaystack, and Proposed Invoice Haystack benchmarks.
Zero-shot results use full document context; entries marked ``--'' reflect context-window limitations.
For V-RAG and VL-RAG, the best accuracy across Top-1/3/5 is reported.
Best result per column is in {bold}.}
\label{tab:vlm_comparison_filtered}
\resizebox{\textwidth}{!}{%
\begin{tabular}{l l c c c c c c c c c}
\toprule
 & & \multicolumn{3}{c}{\textbf{Doc}} & \multicolumn{3}{c}{\textbf{Info}} & \multicolumn{3}{c}{\textbf{Invoice}} \\
\cmidrule(lr){3-5} \cmidrule(lr){6-8} \cmidrule(lr){9-11}
\textbf{Approach} & \textbf{Model} & 100 & 200 & 1000 & 100 & 200 & 1000 & 500 & 1000 & 1500 \\
\midrule
\multirow{4}{*}{Zero-Shot}
 & Gemini-3-Flash  & 74.3 & \textbf{96.3} & -- & \textbf{83.9} & \textbf{87.7} & -- & 18.0 & -- & -- \\
 & GPT-5.2         & 45.0 & 44.0 & -- & 40.6 & 36.1 & -- & 33.0 & -- & -- \\
 & InternVL3-8B    & 78.9 & 75.2 & -- & 55.5 & 56.1 & -- & --   & -- & -- \\
 & Qwen3-VL-8B     & \textbf{94.5} & 93.6 & -- & 72.9 & 74.2 & -- & --   & -- & -- \\
\midrule
\multirow{4}{*}{Model$+$V-RAG}
 & Gemini-3-Flash  & 84.4 & 85.3 & 75.2 & 83.2 & 80.0 & \textbf{74.2} & 51.0 & 48.5 & 44.0 \\
 & GPT-5.2          & 78.9 & 77.8 & 72.5 & 73.5 & 72.9 & 67.1 & 49.5 & 45.5 & 41.5 \\
 & InternVL3-8B    & 71.6 & 66.9 & 60.6 & 55.5 & 54.8 & 51.6 & 32.0 & 30.0 & 26.5 \\
 & Qwen3-VL-8B     & 86.2 & 80.7 & 61.9 & 68.4 & 71.6 & 61.9 & 48.0 & 45.0 & 41.0 \\
\midrule
\multirow{4}{*}{\textbf{Model$+ $VL-RAG}} 
 & Gemini-3-Flash  & 86.2 & 84.4 & \textbf{76.1} & 82.6 & 83.9 & 73.5 & 60.0 & \textbf{56.5} & \textbf{53.0} \\
 & GPT-5.2          & 83.5 & 77.1 & 73.4 & 75.5 & 72.9 & 64.5 & 61.0 & 54.0 & 52.0 \\
 & InternVL3-8B    & 73.4 & 66.9 & 65.1 & 56.1 & 56.1 & 51.6 & 42.5 & 36.0 & 33.5 \\
 & Qwen3-VL-8B     & 87.2   & 83.5   & 73.4  & 72.9   & 73.5   & \textbf{74.2}   & \textbf{63.5} & 56.0 & 52.5 \\
\bottomrule
\end{tabular}}

\end{table*}

\subsection{Ablation Study}

\begin{table*}[tb]
\centering
\caption{Ablation study for VL-RAG on Document Haystack-1000, InfoHaystack-1000, and Invoice Haystack-1500. Results reported as Recall@1 / Recall@3 / Recall@5. Avg. denotes the mean across all three benchmarks.}
\label{tab:ablation}
\resizebox{\textwidth}{!}{%
\begin{tabular}{ccccc|ccc|ccc|ccc|ccc}
\hline
\multicolumn{5}{c|}{\textbf{Components}} & \multicolumn{3}{c|}{\textbf{Doc-1000}} & \multicolumn{3}{c|}{\textbf{Info-1000}} & \multicolumn{3}{c|}{\textbf{Invoice-1500}} & \multicolumn{3}{c}{\textbf{Average}} \\
\textbf{CLIP} & \textbf{SigLIP} & \textbf{OpenCLIP} & \textbf{Text} & \textbf{VLM} & R@1 & R@3 & R@5 & R@1 & R@3 & R@5 & R@1 & R@3 & R@5 & R@1 & R@3 & R@5 \\
\hline
\checkmark & \textcolor{red}{\ding{55}} & \textcolor{red}{\ding{55}} & \textcolor{red}{\ding{55}} & \textcolor{red}{\ding{55}} & 27.5 & 41.3 & 47.7 & 53.6 & 68.4 & 72.3 & 7.0  & 11.0 & 14.0 & 29.4 & 40.2 & 44.7 \\
\textcolor{red}{\ding{55}} & \checkmark & \textcolor{red}{\ding{55}} & \textcolor{red}{\ding{55}} & \textcolor{red}{\ding{55}} & 41.3 & 51.4 & 54.1 & 45.8 & 62.6 & 69.7 & 20.0 & 30.0 & 37.0 & 35.7 & 48.0 & 53.6 \\
\textcolor{red}{\ding{55}} & \textcolor{red}{\ding{55}} & \checkmark & \textcolor{red}{\ding{55}} & \textcolor{red}{\ding{55}} & 41.3 & 59.6 & 66.1 & 57.4 & 72.3 & 79.4 & 19.5 & 24.0 & 28.0 & 39.4 & 52.0 & 57.8 \\
\textcolor{red}{\ding{55}} & \textcolor{red}{\ding{55}} & \textcolor{red}{\ding{55}} & \checkmark & \textcolor{red}{\ding{55}} & 58.7 & 76.1 & 82.6 & 67.1 & 80.7 & 85.2 & 31.5 & 42.5 & 49.5 & 52.4 & 66.4 & 72.4 \\
\checkmark & \checkmark & \textcolor{red}{\ding{55}} & \textcolor{red}{\ding{55}} & \textcolor{red}{\ding{55}} & 42.2 & 58.7 & 63.3 & 58.7 & 74.8 & 79.4 & 14.5 & 23.0 & 29.0 & 38.5 & 52.2 & 57.2 \\
\checkmark & \checkmark & \checkmark & \textcolor{red}{\ding{55}} & \textcolor{red}{\ding{55}} & 45.9 & 66.1 & 74.3 & 65.2 & 76.1 & 81.9 & 21.0 & 30.5 & 35.5 & 44.0 & 57.6 & 63.9 \\
\checkmark & \checkmark & \checkmark & \checkmark & \textcolor{red}{\ding{55}} & 57.8 & 77.1 & 78.0 & 72.3 & 81.3 & 84.5 & 33.5 & 42.0 & 47.0 & 54.5 & 66.8 & 69.8 \\
\textcolor{red}{\ding{55}} & \checkmark & \checkmark & \checkmark & \textcolor{red}{\ding{55}} & 62.4 & 78.9 & 78.9 & 70.3 & 81.3 & 85.8 & 35.0 & 45.0 & 48.5 & 55.9 & 68.4 & 71.1 \\
\checkmark & \checkmark & \checkmark & \checkmark & \checkmark & 77.1 & 82.6 & 82.6 & 83.2 & 85.8 & 86.5 & 48.0 & 52.0 & 52.5 & 69.4 & 73.5 & 73.8 \\
\textcolor{red}{\ding{55}} & \checkmark & \checkmark & \checkmark & \checkmark & 77.1 & 81.7 & 81.7 & 84.5 & 87.7 & 89.0 & 50.0 & 53.5 & 54.0 & \textbf{70.5} & \textbf{74.3} & \textbf{74.9} \\
\hline
\end{tabular}}

\end{table*}

To isolate the contribution of each architectural component, we conduct a systematic ablation on Document Haystack-1000, InfoHaystack-1000, and Invoice Haystack-1500, reporting averages across all three with different encoder combinations (Table~\ref{tab:ablation}).

\smallskip\noindent\textbf{Individual encoder performance.} Vision-only encoders in isolation perform poorly on Invoice Haystack. CLIP alone yields 7.0\% Recall@1, SigLIP 20.0\%, and OpenCLIP 19.5\%, confirming that no single vision encoder provides sufficient discriminative power under visual homogeneity. The text embedding (BGE) alone achieves 31.5\% Recall@1 on Invoice Haystack-1500, with an average Recall@1 of 52.4\% across all benchmarks, substantially outperforming all individual vision encoders in the average.

\smallskip\noindent\textbf{The No-CLIP finding.} A surprising result emerges when examining CLIP's contribution to the ensemble. Adding CLIP to the vision stream consistently degrades performance. The configuration CLIP + SigLIP + OpenCLIP achieves an average Recall@1 of 44.0\%, while SigLIP + OpenCLIP alone (combined with the text stream) performs better. When CLIP is included in the full VL-RAG ensemble (CLIP + SigLIP + OpenCLIP + Text), average Recall@1 drops to 54.5\% compared to the No-CLIP equivalent. On Invoice Haystack-1500 specifically, adding CLIP reduces Recall@1 from 50.0\% to 33.5\%. This counterintuitive result suggests that CLIP's natural-image pre-training introduces retrieval noise, prioritizing global aesthetic similarity over document-level semantic content. 

\smallskip\noindent\textbf{VLM filter contribution.} Adding the VLM filter consistently improves performance across all configurations. The full VL-RAG pipeline (SigLIP + OpenCLIP + Text + VLM Filter) achieves the best average Recall@1 of 70.5\%, compared to 55.9\% without the filter for the same encoder set. On Invoice Haystack-1500, the VLM filter raises Recall@1 from 35.0\% to 50.0\%, with the most notable gains in Recall@3 and Recall@5 (54.0\% and 54.5\% respectively). The filter is particularly effective in homogeneous settings where vector similarity fails to separate the correct document from visually near-identical candidates.

\begin{figure}[tb]
    \centering
    \includegraphics[width=1\linewidth]{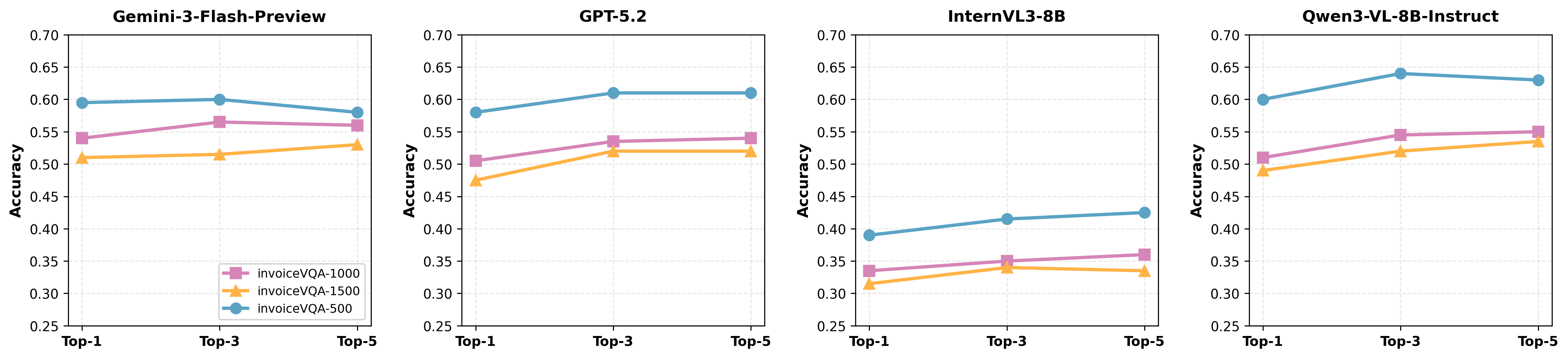}
    \caption{Top-$k$ selection ablation analysis for VQA on our Invoice Haystack benchmark. We evaluate the impact of different retrieval depths ($k=1, 3, 5$) on VQA accuracy across all three corpus scales (Invoice Haystack-500, Invoice Haystack-1000, Invoice Haystack-1500) for four VLMs: Gemini-3-Flash-Preview, GPT-5.2, InternVL3-8B, and Qwen3-VL-8B-Instruct, all integrated with our VL-RAG framework.
}
\label{fig:vl_rag_top_k}
\end{figure}

\smallskip\noindent\textbf{Top-$k$ Retrieval Analysis.} Figure~\ref{fig:vl_rag_top_k} analyses the effect of retrieval depth on VQA accuracy across all four VLMs on our Invoice Haystack benchmark. Generally, accuracy increases as $k$ grows from 1 to 5, since providing more candidate documents raises the likelihood that the correct invoice appears in the VLM context. However, this trend is not monotonic for all models: Gemini and Qwen3-VL both show a slight accuracy decline at top-5 compared to Top-3 on Invoice Haystack-500 (Gemini: 60.0\% $\to$ 58.0\%; Qwen3-VL: 64.0\% $\to$ 63.0\%), suggesting that beyond a certain retrieval depth, additional candidates introduce distractors that degrade answer precision for stronger models. GPT-5.2 and InternVL3 benefit more consistently from increasing $k$ across all corpus scales. Invoice Haystack-500 yields the highest accuracy across all models and $k$ settings, while Invoice Haystack-1500 remains the hardest split, reflecting the greater retrieval difficulty at larger corpus scales. The persistent performance gap across corpus scales, regardless of $k$ confirms that the primary bottleneck is retrieval precision under visual homogeneity rather than the number of candidates passed to the VLM.

\section{Conclusion}

This work demonstrated that embedding collapse is a fundamental and previously unaddressed flaw in multi-document retrieval benchmarks, and that visually homogeneous enterprise collections represent a qualitatively distinct challenge from the diverse documents on which existing methods were developed. We introduced Invoice Haystack to stress-test retrieval under strong visual homogeneity,  where even the strongest vision ensemble reaches only 40.0\% Recall@1 at scale. \textbf{VL-RAG} addresses this through dual-stream fusion of text and visual signals, achieving 50.0\% Recall@1 on Invoice Haystack-1500 while maintaining consistent gains on heterogeneous benchmarks, confirming hybrid encoding as a consistently beneficial strategy regardless of corpus composition. Current limitations include fixed equal-weight fusion and zero-shot encoders without domain-specific fine-tuning; future work should investigate learnable fusion weights conditioned on corpus statistics, encoder fine-tuning on financial documents, and benchmark extension to contracts and purchase orders.

\section*{Acknowledgements}
Naveed Akhtar is a recipient of the Australian Research Council Discovery Early Career Researcher Award (project \# DE230101058) funded by the Australian Government. This research was also supported by The University of Melbourne’s Research Computing Services and the Petascale Campus Initiative. This work is also partially supported by the Google Research Scholar Program Award.

\bibliographystyle{splncs04}
\bibliography{main}
\end{document}